\documentclass{ieeeaccess}
\usepackage{cite}
\usepackage{amsmath,amssymb,amsfonts}
\usepackage{algorithmic}
\usepackage{array}
\usepackage[caption=false,font=normalsize,labelfont=sf,textfont=sf]{subfig}
\usepackage{textcomp}
\usepackage{stfloats}
\usepackage{url}
\usepackage{ragged2e}
\usepackage{verbatim}
\usepackage{graphicx,color}
\usepackage{xcolor}
\usepackage{balance}
\usepackage{multirow}
\usepackage{booktabs}
\usepackage{tabularx}
\def\BibTeX{{\rm B\kern-.05em{\sc i\kern-.025em b}\kern-.08em
    T\kern-.1667em\lower.7ex\hbox{E}\kern-.125emX}}
\AtBeginDocument{\definecolor{tmlcncolor}{cmyk}{0.93,0.59,0.15,0.02}\definecolor{NavyBlue}{RGB}{0,86,125}}

\def\x{{\mathbf x}}

\begin{document}

\title{Next-Day Wildfire Spread Prediction Using Directional Orthogonal Transform Fusion in U-Net}

\author{\uppercase{Ene Meco}\authorrefmark{1}, \uppercase{Yingyi Luo}\authorrefmark{1}, \uppercase{Emadeldeen Hamdan}\authorrefmark{1}, \uppercase{Adam Watts}\authorrefmark{3}, AND \uppercase{Ahmet Enis Cetin, (Fellow, IEEE)}\authorrefmark{1,2}}

\address[1]{
Department of Electrical and Computer Engineering,
University of Illinois Chicago,
Chicago, IL 60607 USA}

\address[2]{
Discovery Partners Institute,
University of Illinois Urbana-Champaign,
Chicago, IL 60601 USA}

\address[3]{
US Forest Service,
Pacific Wildland Fire Science Laboratory,
Seattle, WA 98103 USA}

\begin{abstract}
Next-day wildfire spread prediction is a safety-critical remote-sensing task that requires accurate segmentation of sparse and highly anisotropic fire-front boundaries under strict computational constraints. This paper introduces a wildfire spread prediction method from multi-modal satellite and environmental data using a U-Net with orthogonal transform layers. The U-Net model uses a Walsh--Hadamard Transform (WHT) and Discrete Cosine Transform (DCT) spectral-fusion backbone with a sparse cone-adapted Shearlet residual pathway designed to capture elongated directional structures associated with fire-front propagation. The WHT and DCT branches provide complementary global and frequency-domain representations with lightweight learned spectral scaling, while the Shearlet branch injects anisotropic multi-directional edge information at selected encoder stages. This design replaces heavy learned attention projections with fixed orthogonal transforms and a compact channel-wise fusion gate, substantially reducing model size and computation. On the WildfireSpreadTS benchmark, the proposed ShearFuse-UNet achieves an F1 score of 0.596 with only 267k parameters, outperforming a 14M-parameter ResNet18-based U-Net while using over 50 times fewer parameters. Compared with the WHT+DCT Fusion-UNet backbone, the Shearlet residual improves recall from 0.619 to 0.632, which is desirable in wildfire response settings where missed fire spread is more costly than false alarms. Experiments on the Google Next-Day Wildfire Spread dataset \cite{huot2022wildfire} further show consistent improvements over corresponding transform-fusion baselines. These results indicate that directional transform-domain inductive biases can improve the accuracy--efficiency trade-off for satellite-based wildfire spread prediction.
\end{abstract}

\begin{keywords}
Discrete cosine transform,
Hadamard transform,
satellite data,
shearlet transform,
spectral attention,
U-Net,
wildfire prediction.
\end{keywords}

\maketitle
\markboth
{E. Meco \headeretal: Next-Day Wildfire Spread Prediction Using Directional Orthogonal Transform Fusion in U-Net}
{E. Meco \headeretal: Next-Day Wildfire Spread Prediction Using Directional Orthogonal Transform Fusion in U-Net}
\section{Introduction} 
\label{sec:intro}

\IEEEPARstart{W}{ildfires} are among the most devastating natural hazards,
causing great economic losses and casualties. In 2024, the United States
experienced more than 64,000 wildfires burning over 8.9 million acres
\cite{nicc2024}. In January 2025, catastrophic wildfires in Southern California
burned over 50,000 acres, destroyed more than 16,000 structures, and caused at
least 29 deaths \cite{calfire2025incidents}. Economic losses grew from \$8.6
billion (2012--2016) to over \$81.6 billion (2017--2021)
\cite{iglesias2022fires}.

With the frequency and severity of wildfires continuing to rise due to climate
change, there is an urgent need for novel prediction methods for effective
mitigation and disaster response
\cite{ostoja2023nca,gunay2012entropy,pan2022deep}. Next-day fire spread
prediction, which estimates the areas likely to burn within the following 24
hours, provides critical information for optimal resource allocation and rapid
emergency response. Traditional simulators range from empirical to deterministic
models \cite{singh2025review}, relying on statistical analysis of historical
data \cite{singh2021land,duff2021wildfire,eden2020fire} or physics-based
numerical simulations \cite{rahman2018forest}. These methods often struggle with
the complex, nonlinear interactions among environmental variables. Deep learning provides opportunities for data-driven approaches capable 
of learning spatiotemporal patterns directly from multi-modal observations 
\cite{digiuseppe2025fire}, and has been applied to wildfire problems across 
a range of modalities, from signal and image processing approaches to fire 
detection \cite{cetin2016methods,hong2024wildfire,hamdan2026gwfp} to satellite-based 
next-day spread prediction, the focus of this work.

Earlier CNN-based studies \cite{hodges2019wildland,kantarcioglu2023fire}
predicted fire occurrence from GIS-based remote sensing data but faced challenges
of high computational cost and limited data availability
\cite{Radke2019FireCastLD}. More recent work has employed CNN autoencoders
\cite{huot2022wildfire,shadrin2024wildfire,dong2017learningdeeprepresentationsusing}
and U-Net variants
\cite{gerard2023wildfirespreadts,ronneberger2015unetconvolutionalnetworksbiomedical,zhou2018unetnestedunetarchitecture}
for segmentation-based fire spread prediction, but most efforts have
prioritized dataset collection over architectural efficiency.

Transform-domain methods offer a principled alternative \cite{pan2024multichannel,pan2021fast}. Orthogonal transforms
such as the Hadamard Transform (HT) and DCT establish latent spaces with
decorrelated coefficients, enabling compact and efficient feature learning,
particularly beneficial when training data are limited
\cite{zhu2024probabilistichadamardunetmri,agaian2006hadamard}. However, these
isotropic transforms do not explicitly encode directional structures. Wildfire
spread fronts are strongly anisotropic: they propagate along directional
boundaries shaped by wind, terrain, and vegetation. Shearlet transforms
\cite{kutyniok2012shearlets} address this limitation by providing an optimally
sparse, directionally sensitive multi-scale decomposition.

We propose \textbf{ShearFuse-UNet}, a single unified model that fuses all three
transforms inside a lightweight U-Net encoder. Our contributions are:
\begin{itemize}
  \item A novel triple-branch encoder block combining WHT, DCT, and a
        cone-adapted digital Shearlet residual, with a learned SpectralFusion
        gate, a design loosely inspired by the separation of representation and
        content aggregation used in self-attention, but implemented entirely
        through fixed transform-domain operators rather than learned token
        projections.
  \item A pure-PyTorch implementation of the cone-adapted digital Shearlet
        transform using FFT-based circular convolution, requiring no external
        libraries.
  \item State-of-the-art lightweight performance on WildfireSpreadTS:
        F1\,=\,0.596 (IoU\,=\,0.424) with only 267k parameters,
        outperforming a 14M-parameter ResNet18-based U-Net.
\end{itemize}

\section{ShearFuse-UNet}
\label{sec:method}

\subsection{Overall Architecture}
\label{ssec:overview}

ShearFuse-UNet follows a standard U-Net topology with four encoder stages
(each a MaxPool followed by a DoubleConvWHT block), a bottleneck DoubleConvWHT
block, four symmetric decoder stages (bilinear upsampling, skip-connection
concatenation, double convolution), and a $1{\times}1$ output convolution. The
base channel width is set to $c_1 = 8$, giving channel progression
$8 \to 16 \to 32 \to 64 \to 128$ through the encoder.

\begin{figure*}[t]
\centering
\includegraphics[width=0.75\linewidth]{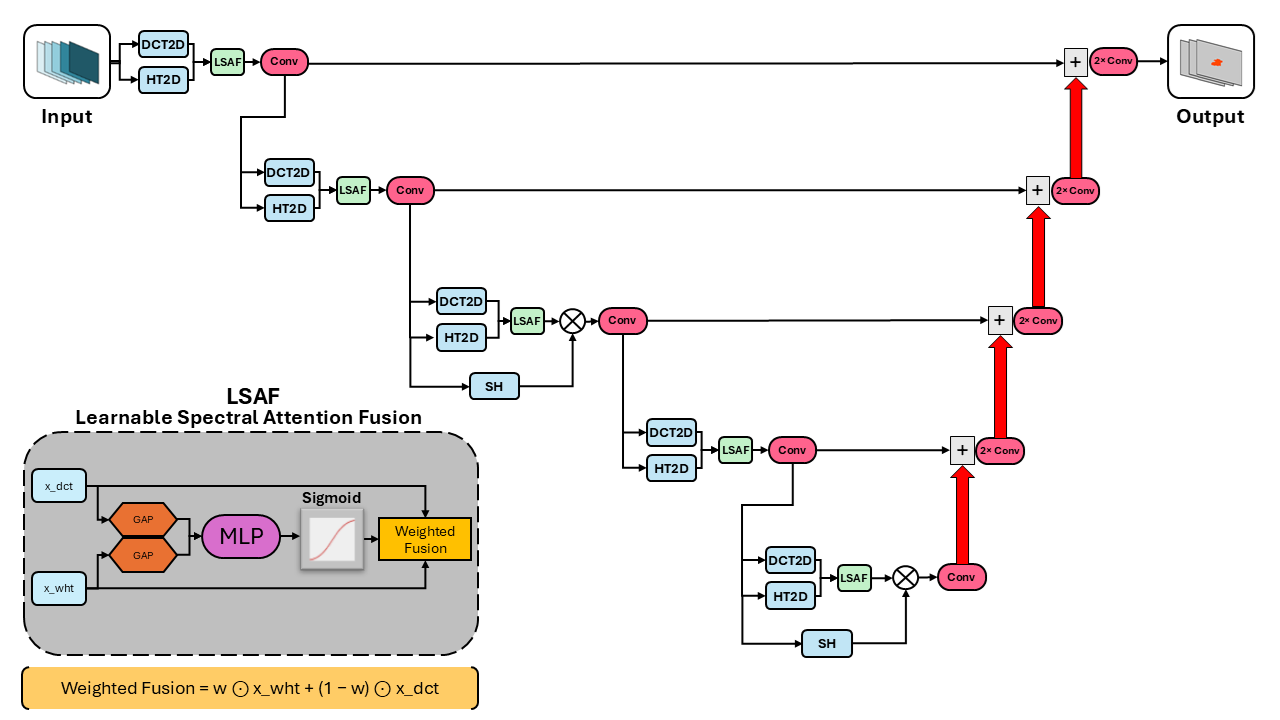}
\caption{Overview of ShearFuse-UNet. Each encoder block contains parallel WHT,
DCT, and optional Shearlet branches. WHT and DCT responses are combined through
SpectralFusion gating, while the Shearlet reconstruction is injected as a
residual pathway. Decoder blocks follow a standard U-Net design with skip
connections.}
\label{fig:arch}
\end{figure*}

What distinguishes each DoubleConvWHT encoder block from a standard double
convolution is the insertion of three complementary transform-domain branches
before the spatial convolutions, as illustrated in Fig.~\ref{fig:arch}. These
branches operate on the input feature map $\mathbf{x} \in
\mathbb{R}^{B \times C \times H \times W}$ and produce a spectrally filtered
output that replaces the first convolution's input. All transform sizes are
derived automatically from the spatial resolution at each encoder stage:
$(H, W)$, $(H/2, W/2)$, $(H/4, W/4)$, $(H/8, W/8)$, and $(H/16, W/16)$.

\subsection{Walsh--Hadamard Transform Branch}
\label{ssec:wht}

The WHT branch applies a 2D Fast Walsh--Hadamard Transform along both spatial
dimensions using a butterfly algorithm. Let $\mathbf{H}_N$ be the
(unnormalized) Hadamard matrix of order $N$, satisfying
$\mathbf{H}_N \mathbf{H}_N^{\top} = N\mathbf{I}$. For each channel
$c = 1,\ldots,C$ the 2D WHT is
\begin{equation}
  \widehat{\mathbf{X}}_c = \mathbf{H}_H \, \mathbf{X}_c \, \mathbf{H}_W,
  \label{eq:wht2d}
\end{equation}
where $\mathbf{H}_H$ and $\mathbf{H}_W$ are the Hadamard matrices of the
respective spatial dimensions, optionally reordered into Walsh (sequency)
ordering \cite{agaian2006hadamard}. A learnable spectral scaling matrix
$\mathbf{V}_{\mathrm{WHT}} \in \mathbb{R}^{H \times W}$ is applied
element-wise:
\begin{equation}
  E_c(i,j) = V_{\mathrm{WHT}}(i,j)\,\widehat{X}_c(i,j),
  \label{eq:wht_scale}
\end{equation}

\begin{figure}[t]
\centering
\includegraphics[width=0.25\linewidth]{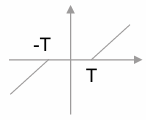}
\caption{Soft-thresholding nonlinearity used in all three spectral branches.}
\label{fig:threshold}
\end{figure}

followed by soft-thresholding \cite{donoho1995denoising} with fixed per-coefficient thresholds
$\mathbf{T}_{\mathrm{WHT}} \in \mathbb{R}^{H \times W}_{\geq 0}$, illustrated in Fig.~\ref{fig:threshold},
initialized from $\mathcal{U}(0, 0.01)$:
\begin{equation}
  S_T(e) = \operatorname{sgn}(e)\,\max(|e| - T,\, 0).
  \label{eq:soft_thresh}
\end{equation}
Reconstruction uses the inverse WHT:
\begin{equation}
  \mathbf{Y}^{\mathrm{WHT}}_c
  = \frac{1}{HW}\,\mathbf{H}_H\,\mathbf{Z}^{\mathrm{WHT}}_c\,\mathbf{H}_W.
  \label{eq:iwht}
\end{equation}

\subsection{Discrete Cosine Transform Branch}
\label{ssec:dct}

The DCT branch applies a 2D orthonormal DCT-II to each channel. The basis
matrix $\mathbf{D}_N \in \mathbb{R}^{N \times N}$ satisfies
$\mathbf{D}_N \mathbf{D}_N^{\top} = \mathbf{I}$ and is constructed as
\begin{equation}
  D_N(k,n) = \alpha_k \cos\!\left(\frac{\pi(n+\tfrac{1}{2})k}{N}\right),
  \label{eq:dct_matrix}
\end{equation}
where $\alpha_0 = 1/\sqrt{N}$ and $\alpha_k = \sqrt{2/N}$ for $k > 0$. The
2D forward transform is
\begin{equation}
  \widetilde{\mathbf{X}}_c = \mathbf{D}_H\,\mathbf{X}_c\,\mathbf{D}_W^{\top}.
  \label{eq:dct2d}
\end{equation}
An optional spectral compression retains only the top-$r$ fraction of
low-frequency coefficients, where $r \in (0,1]$ is the compression ratio.
A learnable scaling matrix $\mathbf{V}_{\mathrm{DCT}}$ and fixed soft-threshold
$\mathbf{T}_{\mathrm{DCT}}$, illustrated in Fig.~\ref{fig:threshold}, initialized from $\mathcal{U}(0, 0.01)$,
are applied to the retained coefficients. The
inverse DCT-II reconstructs the spatial features:
\begin{equation}
  \mathbf{Y}^{\mathrm{DCT}}_c
  = \mathbf{D}_H^{\top}\,\mathbf{Z}^{\mathrm{DCT}}_c\,\mathbf{D}_W.
  \label{eq:idct}
\end{equation}

\subsection{Shearlet Transform: Background and Motivation}
\label{sec:shearlet_background}
Classical wavelet transforms provide a powerful framework for multi-scale signal analysis, but they suffer from a fundamental limitation when applied to two-dimensional data: isotropy. A separable 2D wavelet basis decomposes an image into subbands at multiple scales but treats all spatial directions identically, offering only a coarse $\{0^\circ, 45^\circ, 90^\circ\}$ directional vocabulary inherited from the tensor-product construction. This is adequate for isotropic textures but poorly suited to signals dominated by elongated, directionally coherent structures,  precisely the geometry of wildfire spread fronts, which propagate along anisotropic boundaries shaped by wind vectors, terrain slope, and vegetation corridors.

The Shearlet transform \cite{kutyniok2012shearlets} was developed to overcome this limitation. It belongs to the broader class of \emph{affine-like systems} that generate a frame for $L^2(\mathbb{R}^2)$ from a single generating function $\psi \in L^2(\mathbb{R}^2)$ through a structured family of geometric transformations. The key innovation is the replacement of the rotation group, which acts on a circle and is incompatible with the Cartesian lattice of digital images, with the \emph{shearing group}, whose action is expressed by integer-valued shearing matrices and is therefore natively compatible with discrete rectangular grids.

Concretely, a shearlet system is generated by composing three operations on the generator $\psi$:

\begin{itemize}
    \item \textbf{Anisotropic dilation} by the parabolic scaling matrix $A_a = \mathrm{diag}(a,\, a^{1/2})$ for scale parameter $a > 0$, which stretches the support parabolically rather than isotropically. This implements the \emph{parabolic scaling law} $\mathrm{width} \sim \mathrm{length}^2$, the hallmark of anisotropic singularities such as curve discontinuities and sharp edges.

    \item \textbf{Shearing} by the matrix $S_k = \bigl[\begin{smallmatrix}1 & k \\ 0 & 1\end{smallmatrix}\bigr]$ for shear index $k \in \mathbb{Z}$, which tilts the support of $\psi$ to a prescribed orientation without distorting area, producing directional selectivity at each angle.

    \item \textbf{Translation} by $t \in \mathbb{R}^2$, which localises the resulting element in the spatial domain.
\end{itemize}

The continuous shearlet system is then the collection $\{\psi_{a,k,t}\}$ defined in the frequency domain as
\begin{equation}
    \hat{\psi}_{a,k,t}(\xi) = a^{3/4}\,\hat{\psi}\!\left(A_a^{-1} S_k^{-\top} \xi\right) e^{-2\pi i \xi \cdot t},
    \label{eq:shearlet_def}
\end{equation}
where the $a^{3/4}$ normalization ensures that $\{\psi_{a,k,t}\}$ forms a tight frame, a stable, redundant system that guarantees perfect reconstruction and preserves inner-product norms up to a global constant \cite{kutyniok2012shearlets}.

The fundamental theoretical virtue of the Shearlet transform is its \emph{optimal sparsity} for cartoon-like images, i.e., piecewise-smooth functions $f \in L^2(\mathbb{R}^2)$ with discontinuities along $C^2$ curves. This property is directly relevant to wildfire spread prediction: the
next-day fire masks used in this work are binary segmentation maps whose
boundaries trace the fire-front perimeter, a geometry that closely resembles
the cartoon-image model of piecewise-smooth regions separated by smooth curves.
The Shearlet transform is therefore a principled and application-motivated
choice for encoding fire-front structure. Under this model, the $n$-term 
nonlinear approximation error of the Shearlet expansion satisfies
\begin{equation}
    \|f - f_n\|^2_{L^2} = O\!\left(n^{-2}(\log n)^3\right), \quad n \to \infty,
    \label{eq:shearlet_approx}
\end{equation}
which matches the information-theoretic optimum achievable by any 
representation system \cite{kutyniok2012shearlets} and strictly outperforms 
the $O(n^{-1})$ rate of separable wavelets. For the binary fire masks used 
in this work, where $f = \mathbf{1}_B$ and the approximation error depends 
solely on the $C^2$ regularity of the fire-front perimeter $\partial B$, 
this rate specialises to
\begin{equation}
    \|\mathbf{1}_B - (\mathbf{1}_B)_n\|^2_{L^2} 
    = O\!\left(n^{-2}(\log n)^3\right), \quad n \to \infty,
    \label{eq:shearlet_binary}
\end{equation}
where the implicit constant depends only on the curvature of $\partial B$
and not on interior image content. Intuitively, this is because parabolic 
dilation aligns the long axis of each shearlet element with the local tangent 
of the discontinuity curve, allowing a single large coefficient to represent 
a long, smooth segment of an edge rather than requiring many isotropic atoms 
to approximate it.

For digital implementation, the continuous construction must be discretised and made compatible with finite rectangular grids. The \emph{cone-adapted} discrete Shearlet transform \cite{kutyniok2012shearlets} partitions the 2D frequency plane $\hat{\mathbb{R}}^2$ into two cones,
\begin{align}
    \mathcal{C}_h &= \bigl\{(\xi_1,\xi_2) : |\xi_2/\xi_1| \leq 1\bigr\}, \\
    \mathcal{C}_v &= \bigl\{(\xi_1,\xi_2) : |\xi_1/\xi_2| < 1\bigr\},
\end{align}
plus a low-frequency region near the origin. Separate shearlet filter banks are constructed within each cone, ensuring that all directions in $[0^\circ, 180^\circ)$ receive uniform directional coverage without the boundary artifacts that arise when a single directional system is applied naively to the full frequency plane. This is the construction adopted in Section~\ref{sec:shearlet_branch}, implemented entirely in PyTorch via FFT-based circular convolution without external libraries.

The practical consequence for wildfire spread prediction is twofold. First, the Shearlet subbands at each scale $s$ and shear index $k$ respond selectively to edges and boundaries oriented in the direction encoded by $k$, providing an explicit geometric representation of fire-front structure that isotropic transforms, including both the WHT and DCT branches described below, are intrinsically unable to capture. Second, because the Shearlet frame is \emph{redundant} rather than orthonormal, reconstruction is stable even when individual subband coefficients are suppressed by soft-thresholding or learnable gating, making the transform a natural substrate for the residual refinement pathway introduced in Section~\ref{sec:shearlet_branch}.

\subsection{SpectralFusion Gate}
\label{ssec:fusion}

The WHT and DCT outputs $\mathbf{Y}^{\mathrm{WHT}}$ and
$\mathbf{Y}^{\mathrm{DCT}}$ are combined by a channel-wise SpectralFusion
module. Global average pooling extracts per-channel statistics:
\begin{equation}
  g_1 = \mathrm{GAP}(\mathbf{Y}^{\mathrm{WHT}}),\quad
  g_2 = \mathrm{GAP}(\mathbf{Y}^{\mathrm{DCT}}),
\end{equation}
which are concatenated and passed through a two-layer MLP with ReLU and sigmoid
activations to produce a channel gate $\mathbf{w} \in (0,1)^C$:
\begin{equation}
  \mathbf{F}_{\mathrm{fused}}
  = \mathbf{w} \odot \mathbf{Y}^{\mathrm{WHT}}
  + (1-\mathbf{w}) \odot \mathbf{Y}^{\mathrm{DCT}},
  \label{eq:fusion}
\end{equation}
where $\odot$ denotes broadcast element-wise multiplication. The hidden
dimension of the MLP is $\max(4, C/r)$ with reduction ratio $r=8$.

\subsection{Relationship to transformer Self-Attention}

The critical distinction of ShearFuse-UNet from transformer-based
architectures is the replacement of the three learned projection
matrices $\mathbf{W}^Q$, $\mathbf{W}^K$, $\mathbf{W}^V$ with
mathematically fixed operators, the Hadamard matrix $\mathbf{H}_N$,
the DCT-II basis $\mathbf{D}_N$, and the Shearlet filter bank
$\{\psi_{s,k}\}$. This substitution reduces the model to 267k
parameters while encoding inductive biases analytically that a
transformer must discover from data: global spatial mixing, frequency
selectivity, and directional sensitivity. Only the per-coefficient
scaling weights and subband gains are learned. We present the QKV
correspondence below as a motivational framework for understanding how
the three branches divide their computational roles, not as a claim of
mathematical equivalence.

In canonical self-attention, given an input sequence, the Query~($Q$)
and Key~($K$) projections interact to produce a soft attention map over
the input, while the Value~($V$) projection carries the content that is
modulated by that map:
\begin{equation}
    \mathrm{Attention}(Q,K,V)
    = \mathrm{softmax}\!\left(\frac{QK^{\top}}{\sqrt{d}}\right)V.
    \label{eq:attention}
\end{equation}
where the softmax normalization is defined as
\begin{equation}
    \mathrm{softmax}(\mathbf{z})_i
    = \frac{e^{z_i}}{\sum_{j} e^{z_j}},
    \label{eq:softmax}
\end{equation}
The role of $Q$ and $K$ is to determine which content is relevant; the
role of $V$ is to provide what that content actually is.

In ShearFuse-UNet, the same functional decomposition emerges from the
structure of the three branches:

\begin{itemize}

\item \textbf{WHT $\leftrightarrow$ Query.} The Hadamard transform
projects the feature map into a flat, globally mixed sequency domain
where every output coefficient depends on every spatial input. The
learnable scaling $\mathbf{V}_{\mathrm{WHT}}$ and thresholding act as a
structured, sparse mask over sequency coefficients, determining which
global mixing patterns are retained and amplified. Like a query
projection, $\mathbf{H}_N$ is a fixed orthogonal operator; it selects
which representations are relevant without itself carrying content.

\item \textbf{DCT $\leftrightarrow$ Key.} The DCT projects the same
input into an energy-compacted frequency domain where low-frequency
coefficients capture smooth, slowly varying spatial patterns and
high-frequency coefficients encode fine detail. Like a key projection,
it provides a complementary decomposition of the same input against
which the WHT representation is compared. The SpectralFusion gate
compares the two spectral representations channel-by-channel via global
average pooling and an MLP, producing a soft weighting $w$:
\begin{equation}
    w = \sigma\!\left(\mathrm{MLP}([g_1;\,g_2])\right).
    \label{eq:gate}
\end{equation}
This gate is functionally analogous to the attention map
$\mathrm{softmax}(QK^{\top}/\!\sqrt{d})$ in that both produce a
data-dependent channel weighting from two complementary
representations, though it operates on globally pooled statistics rather
than spatial token interactions and produces independent per-channel
sigmoid weights rather than a competitive softmax.

\item \textbf{Shearlet $\leftrightarrow$ Value.} The Shearlet branch
most closely parallels the Value stream. It does not participate in the
gating computation at all; instead it independently decomposes the
input into directional, anisotropic subbands and reconstructs a
filtered version $\hat{x}_{\mathrm{sh}}$ that is injected as a
residual after the WHT/DCT fusion. Like the Value projection, the
Shearlet branch provides \emph{what to return} rather than determining
\emph{where to look}: it contributes directionally selective edge
responses along fire-front boundaries that the isotropic WHT and DCT
representations intrinsically cannot encode. Unlike the Value stream in
standard attention, however, the Shearlet output is not modulated by
the gate $w$, it enters as an additive residual rather than as a
content stream weighted by the attention map. This is a deliberate
design choice: the residual formulation allows the Shearlet branch to
contribute independently of the spectral gating, and its learnable
subband gains $\{v_i\}$ can decay toward zero if the directional signal
provides no benefit, without disturbing the WHT/DCT fusion pathway.

\end{itemize}

Table~\ref{tab:qkv} summarises this correspondence; we stress that it
is functional rather than mathematical.

\begin{table*}[t]
\centering
\caption{Conceptual correspondence between self-attention and the
proposed encoder block. The mapping is functional rather than
mathematical.}
\label{tab:qkv}
\begin{tabular}{lll}
\toprule
Transformer & ShearFuse-UNet & Key difference \\
\midrule
Query $Q$            & WHT branch           & Fixed orthogonal operator vs.\ learned projection \\
Key $K$              & DCT branch           & Fixed cosine basis vs.\ learned projection         \\
$QK^{\top}$          & SpectralFusion MLP   & Channel statistics, not spatial token interactions \\
Softmax attention    & Sigmoid gate $w$     & Independent gating vs.\ competitive normalization  \\
Value $V$            & Shearlet branch      & Additive residual, not modulated by gate           \\
Learned $W^Q,W^K,W^V$ & Fixed $\mathbf{H}_N,\mathbf{D}_N,\psi_{s,k}$ & Parameter-free by construction \\
\bottomrule
\end{tabular}
\end{table*}

The critical distinction from standard transformer attention is that
ShearFuse-UNet replaces the three $d \times d$ learned projection
matrices $\mathbf{W}^Q$, $\mathbf{W}^K$, $\mathbf{W}^V$ with
mathematically fixed orthogonal and frame-theoretic operators: the
Hadamard matrix, the DCT-II basis, and the Shearlet filter bank. This
substitution has profound practical consequences. First, the projection
operators are parameter-free by construction (only the per-coefficient
scaling weights are learned), reducing the model to 267k
parameters versus tens of millions in a comparable Vision transformer.
Second, the bases are analytically optimal for their respective tasks:
the WHT maximises global mixing with $\pm 1$ arithmetic, the DCT
minimises energy compaction error for natural signals~\cite{ahmed1974discrete},
and the Shearlet achieves provably optimal sparsity for anisotropic
singularities~\cite{kutyniok2012shearlets}. Third, the block is spatially
equivariant and does not require positional encodings, since the
transforms operate on fixed spatial grids at each resolution.

This is particularly advantageous in the low-data regime of wildfire
spread prediction, where the training set is too small to reliably
learn full-rank attention projections but sufficient to learn the
lightweight spectral gating weights.

\subsection{Cone-Adapted Digital Shearlet Residual Branch}
\label{sec:shearlet_branch}

\subsubsection{Filter Construction}

We build a bank of real-valued shearlet filters entirely in PyTorch, without
external libraries. A Meyer-type smooth window $\nu: [0,1] \to [0,1]$ with
$C^1$ regularity is defined as
\begin{equation}
  \nu(x) = x^2(3-2x).
  \label{eq:meyer}
\end{equation}
For each scale $s \in \{1,\ldots,J\}$ with anisotropic dilation $a = 2^{-s}$
and shear index $k \in \{-\lfloor K/2\rfloor,\ldots,\lfloor K/2\rfloor - 1\}$,
the shearlet filter is constructed in the frequency domain as
\begin{equation}
  \hat{\psi}_{s,k}(\xi,\eta)
  = \psi_r\!\left(\frac{|\xi|}{a^2}\right)
    \cdot \psi_a\!\left(\frac{\eta/|\xi| - k}{1}\right),
  \label{eq:shearlet_freq}
\end{equation}
where the radial window $\psi_r$ selects an annular frequency support and the
angular window $\psi_a$ enforces directional selectivity. A low-pass residual
filter
\begin{equation}
  \hat{\phi}(\xi,\eta) = \nu\!\left(1 - \frac{\sqrt{\xi^2+\eta^2}}{0.25}\right)
  \label{eq:lowpass}
\end{equation}
captures the coarse-scale component. Spatial filters are obtained via the
inverse FFT, yielding $n_f = J \cdot K + 1$ real-valued spatial filters.

\subsubsection{Analysis and Synthesis}

Analysis is performed via FFT-based circular convolution:
\begin{equation}
  \mathcal{SH}(\mathbf{x})_{s,k}
  = \mathcal{F}^{-1}\!\bigl(\mathcal{F}(\mathbf{x})
    \cdot \mathcal{F}(\psi_{s,k})\bigr).
  \label{eq:sh_analysis}
\end{equation}
Synthesis sums responses convolved with spatially-flipped analysis filters:
\begin{equation}
  \hat{\mathbf{x}}_{\mathrm{sh}}
  = \frac{1}{n_f}\sum_{s,k}
    \mathcal{F}^{-1}\!\bigl(\mathcal{F}(\mathcal{SH}(\mathbf{x})_{s,k})
    \cdot \overline{\mathcal{F}(\psi_{s,k})}\bigr).
  \label{eq:sh_synthesis}
\end{equation}

\subsubsection{Learnable Spectral Gating and Residual Integration}

Each shearlet subband is modulated by a learnable per-subband gain
$v_i \in \mathbb{R}$ and a fixed soft threshold $T_i \geq 0$,
initialized from $\mathcal{U}(0, 0.01)$:
\begin{equation}
  \widetilde{\mathcal{SH}}_i
  = S_{T_i}\!\bigl(v_i \cdot \mathcal{SH}(\mathbf{x})_i\bigr).
  \label{eq:sh_gate}
\end{equation}
The reconstructed shearlet output is added to the fused WHT/DCT response as
a residual:
\begin{equation}
  \mathbf{F} = \mathbf{F}_{\mathrm{fused}} + \hat{\mathbf{x}}_{\mathrm{sh}}.
  \label{eq:residual}
\end{equation}
This residual formulation ensures that the Shearlet branch acts as a
modular refinement pathway: if it provides no beneficial directional signal,
its learnable gains can naturally decay toward zero without affecting the
WHT/DCT-based representation. The Shearlet filters are stored as
non-trainable buffers (constructed lazily during the first forward pass),
while only the subband gains $\{v_i\}$ and thresholds $\{T_i\}$ are
learnable.

We further study how frequently the Shearlet residual should be injected
across encoder stages. In an ablation comparing full-stage activation with
a sparser configuration (applying the Shearlet branch at every second
encoder stage), we observe that overly frequent injection slightly degrades
performance, reducing the validation F1 score by approximately $0.5\%$.
This indicates that dense directional augmentation may introduce redundancy
with the isotropic WHT/DCT spectral representations.

Based on this trade-off, we adopt a sparse activation strategy in which the
Shearlet residual is applied only at the \texttt{down2} ($H/4 \times W/4$) and
\texttt{down4} ($H/16 \times W/16$) encoder stages, where mid- and fine-scale
directional structures are most informative for fire-front delineation. The
remaining stages rely solely on the WHT + DCT Spectral Fusion, which helps
maintain a low parameter count while preserving expressive power.

\subsection{Decoder and Output}
\label{ssec:decoder}

The decoder mirrors the encoder with four upsampling stages. Each Up block
bilinearly upsamples the lower-resolution feature map by $2{\times}$, pads it
to match the encoder skip-connection dimensions, concatenates along the channel
axis, and applies a standard DoubleConv (two $3{\times}3$ convolutions, each
followed by BatchNorm and ReLU). The output is produced by a $1{\times}1$
convolution mapping to the desired number of classes.

\section{Experimental Results}
\label{sec:experiments}

\subsection{Datasets}
\label{ssec:datasets}

WildfireSpreadTS \cite{gerard2023wildfirespreadts} provides
multi-temporal satellite observations at $128 \times 128$ spatial resolution
with 40 input channels, representing a challenging and realistic benchmark for
next-day fire spread prediction.

To further validate the generalizability of ShearFuse-UNet, we evaluate it on
the Next-Day Wildfire Spread dataset released by Google Research
\cite{huot2022wildfire} in 2023, a benchmark introduced concurrently with the
WildfireSpreadTS dataset. Each sample consists of $64 \times 64$ patches with
12 input channels derived from satellite imagery and weather data, split into
training, validation, and test sets at an 8:1:1 ratio.

\subsection{Implementation Details}
\label{ssec:impl}

ShearFuse-UNet is implemented in PyTorch with base channel width $c_1 = 8$,
Walsh-ordered WHT, DCT compression ratio $r = 0.7$, $J = 2$ shearlet scales,
$K = 4$ shear directions per scale ($n_f = 9$ filters), and soft thresholding
throughout. The shearlet configuration was selected empirically by evaluating
multiple scale--direction combinations. Specifically, configurations
$(J,K)\in\{(2,4),(3,8)\}$ were compared, where $J$ denotes the number of
scales and $K$ the number of shear directions per scale. The configuration
$J=2,\ K=4$ achieved a slightly higher validation F1 score ($+0.1\%$) while
also reducing the number of directional subbands and computational overhead.
Performance is stable across shearlet configurations, suggesting the benefit
derives from the directional decomposition principle rather than specific filter
bank geometry. $J=2,\ K=4$ was therefore selected for its lower computational
overhead while matching the performance of the denser $J=3,\ K=8$ configuration.

The DCT compression ratio was selected through an empirical ablation study on
the WHT+DCT fusion baseline. Compression ratios of
$r \in \{0.60, 0.65, 0.70, 0.75\}$ were evaluated, where $r$ denotes the
fraction of low-frequency coefficients retained after the DCT transform.
Among the tested settings, $r=0.70$ achieved the highest validation F1 score
and was therefore adopted for all subsequent experiments. All spectral scaling weights are initialized to one. Soft-thresholding is
applied with fixed thresholds initialized from $\mathcal{U}(0, 0.01)$;
an ablation study found that making the thresholds learnable degraded
validation F1, so thresholds are held constant throughout training.
The model has a total of \textbf{267k parameters}.

Training uses the AdamW optimizer with an initial learning rate of
$1 \times 10^{-3}$, linear warmup over 2 epochs, and cosine annealing
decay over the remaining epochs, for a total of 200 epochs. The composite loss is
\begin{equation}
  \mathcal{L}
  = \lambda_{\mathrm{BCE}}\,\mathcal{L}_{\mathrm{BCE}}
  + \lambda_{\mathrm{Dice}}\,\mathcal{L}_{\mathrm{Dice}}
  + \lambda_{\mathrm{Focal}}\,\mathcal{L}_{\mathrm{Focal}},
  \label{eq:loss}
\end{equation}
with $\lambda_{\mathrm{BCE}} = 0.4$, $\lambda_{\mathrm{Dice}} = 0.3$,
$\lambda_{\mathrm{Focal}} = 0.3$. The loss weights were selected via 
a partial ablation over a representative subset of configurations; 
exhaustive search was not performed due to the substantial training 
time required per run.

For the Google dataset experiments, we adopt the customized preprocessing pipeline introduced in \cite{luo2025tdfusion}: uncertain (missing) pixels encoded as $-1$ are remapped to $0$, background pixels are perturbed by values sampled from $\mathcal{U}(0.01, 0.03)$, and fire pixels are replaced by values from $\mathcal{U}(0.80, 0.99)$. A Gaussian-mixture softening step then blurs the pre-fire mask and wind-speed channel with standard deviations $\{\sigma_k\} = \{0.4, 0.8\}$ and averages the resulting maps to produce a soft probability input, attenuating the sparsity of fire pixels and improving boundary generalisation.

\begin{table}[t]
\centering
\caption{Effect of DCT compression ratio on WHT+DCT fusion baseline performance.}
\label{tab:dct_ablation}
\begin{tabular}{|c|c|c|c|}
\hline
\textbf{Compression Ratio} & \textbf{Precision} & \textbf{Recall} & \textbf{F1} \\
\hline
0.60 & 0.523 & 0.675 & 0.590 \\
0.65 & 0.557 & 0.637 & 0.594 \\
\textbf{0.70} & \textbf{0.573} & 0.619 & \textbf{0.595} \\
0.75 & 0.540 & 0.645 & 0.588 \\
\hline
\end{tabular}
\end{table}

\begin{table}[t]
\centering
\caption{Results on the WildfireSpreadTS dataset \cite{gerard2023wildfirespreadts}.
         All models predict next-day fire spread.}
\label{tab:wfts_main}
\renewcommand{\arraystretch}{1.15}
\setlength{\tabcolsep}{6pt}
\begin{tabular}{|l|c|c|c|c|}
\hline
\textbf{Model} & \textbf{Parameters} & \textbf{Precision} & \textbf{Recall} & \textbf{F1} \\
\hline
Baseline (ResNet18)~\cite{gerard2023wildfirespreadts}
  & 14M   & 0.536 & 0.654 & 0.589 \\
\hline
Base Swin-UNet
  & 99M   & 0.564 & 0.623 & 0.592 \\
\hline
Tiny Swin-UNet
  & 34M   & 0.550 & 0.640 & 0.591 \\
\hline
HT-UNet
  & 169k  & 0.510 & 0.646 & 0.570 \\
\hline
Fusion-UNet
  & 267k  & 0.573 & 0.619 & 0.595 \\
\hline
\textbf{ShearFuse-UNet}
  & \textbf{267k} & \textbf{0.564} & \textbf{0.632} & \textbf{0.596} \\
\hline
\end{tabular}
\end{table}

\begin{figure}[b]
\centering
\includegraphics[width=\linewidth]{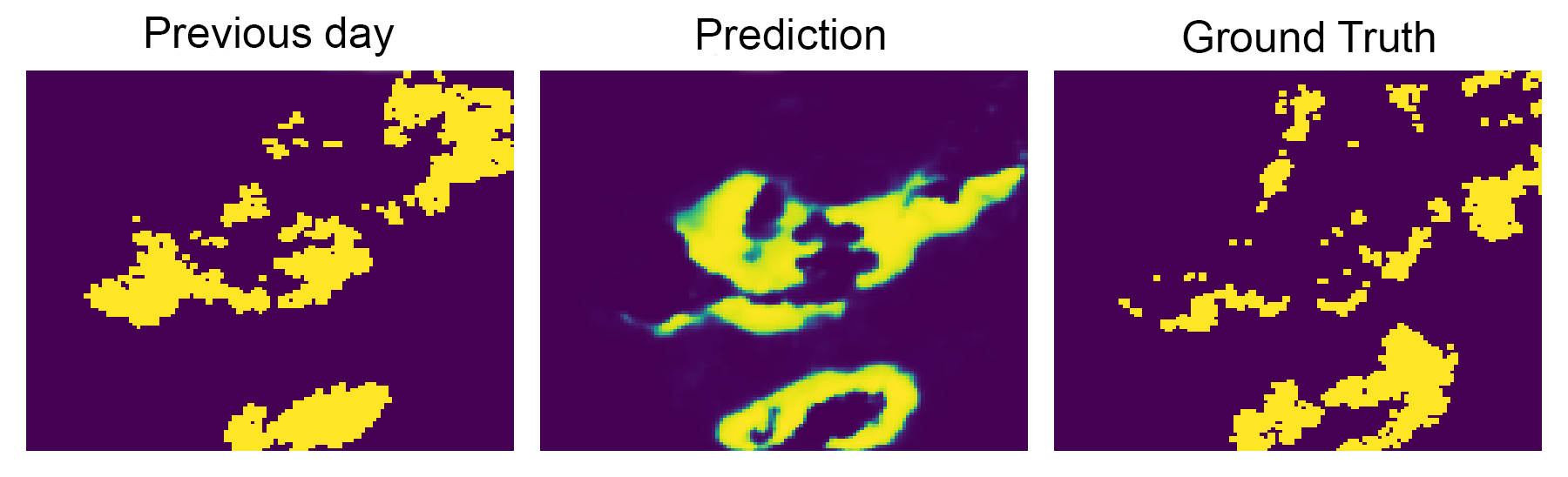}
\includegraphics[width=\linewidth]{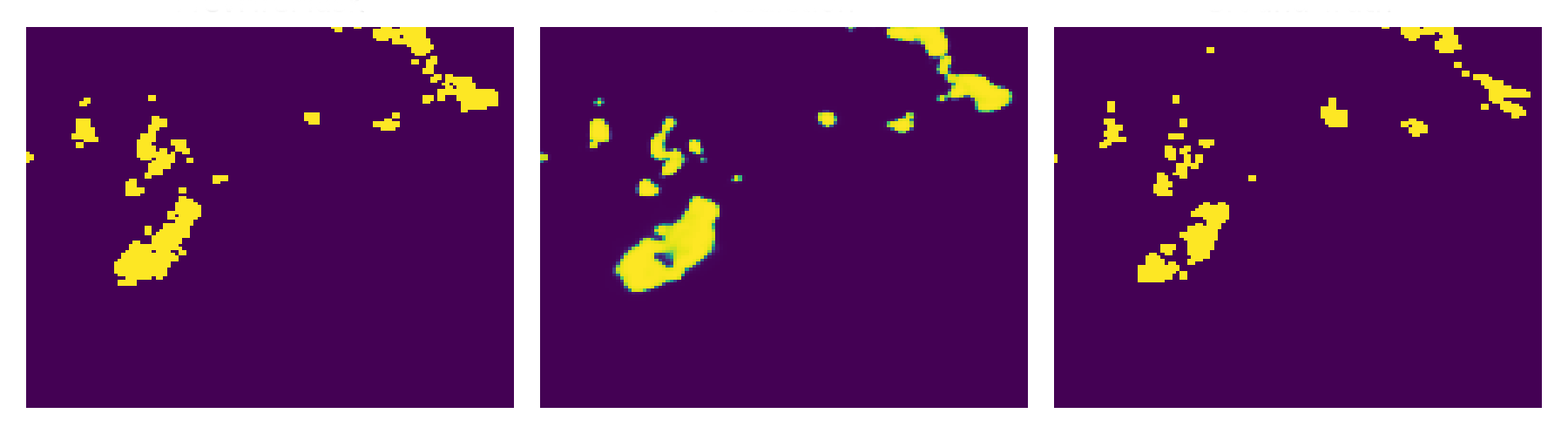}
\includegraphics[width=\linewidth]{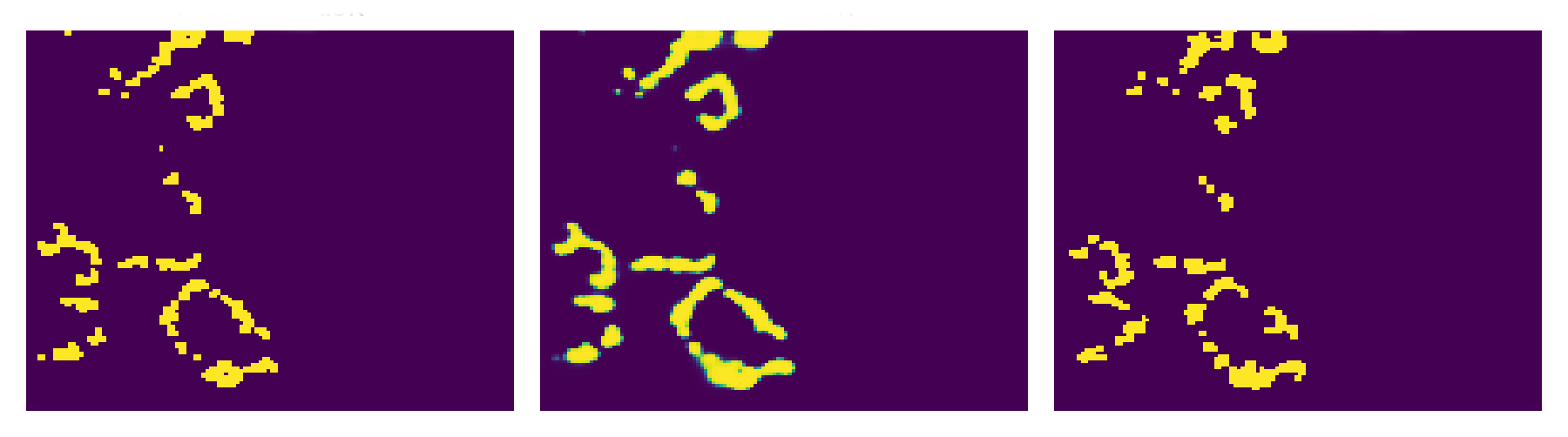}
\caption{Qualitative inference results on three representative WildfireSpreadTS
samples. Each column shows (a) the previous-day fire mask, (b) the next-day
spread prediction, and (c) the ground-truth.}
\label{fig:inference}
\end{figure}

\subsection{Results}
\label{ssec:wfts}

Fig.~\ref{fig:inference} shows representative qualitative results, where
ShearFuse-UNet predicts the overall fire-spread pattern for the following day. Table~\ref{tab:wfts_main} shows that ShearFuse-UNet achieves
the best F1 score of \textbf{0.596} among all lightweight
models. It
surpasses both Swin-UNet variants while requiring up to 370$\times$ fewer
parameters, and outperforms the ResNet18 baseline with a 52$\times$ reduction
in parameter count. Compared with the WHT+DCT Fusion-UNet baseline under the same parameter budget (267k parameters), the addition of the Shearlet residual branch shifts performance toward higher sensitivity, increasing recall from 0.619 to 0.632 while reducing precision from 0.573 to 0.564 and maintaining nearly identical F1 performance (0.595 vs. 0.596). In wildfire emergency response, missed fire detections carry substantially
greater cost than false alarms \cite{conformal2025wildfire}. The
precision--recall profile of ShearFuse-UNet is therefore not merely a side
effect of the Shearlet branch but reflects a deliberate and
application-appropriate design outcome: the directional sensitivity of the
Shearlet filters preferentially recovers elongated fire-front boundaries that
isotropic WHT and DCT representations fail to detect.
Table~\ref{tab:efficiency} reports computational cost and inference latency.
ShearFuse-UNet requires only 1.35 GFLOPs and 3.03 ms per sample, compared to
6.63 GFLOPs and 8.31 ms for the Swin-B UNet, while achieving a higher F1
(0.596 vs.\ 0.592) at one-370th the parameter count.
Table~\ref{tab:wfts_days} further investigates the effect of temporal context on model performance. Extending the input from a single observation day to two consecutive days yields a consistent improvement across all metrics, with F1 increasing from 0.596 to 0.600 and precision from 0.564 to 0.570, at a negligible parameter cost of only 2k additional weights. 

\begin{figure}[t]
\centering
\includegraphics[width=\linewidth]{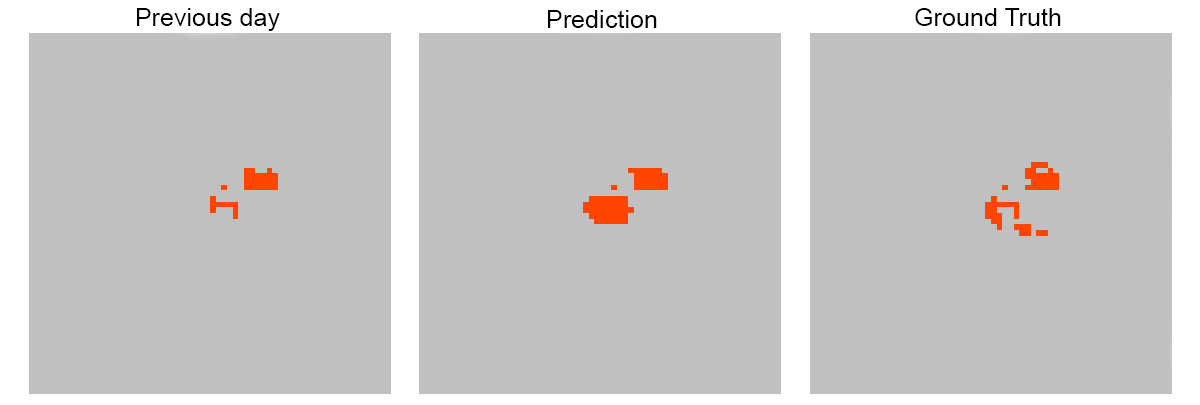}
\includegraphics[width=\linewidth]{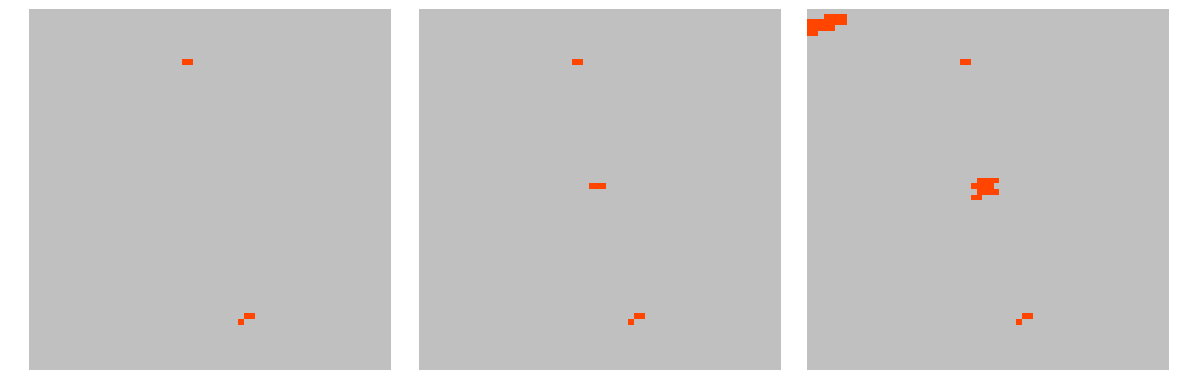}
\caption{Qualitative inference results on two representative Google Next-Day
Wildfire Spread samples. Each row shows (a) the previous-day fire mask,
(b) the next-day spread prediction, and (c) the ground-truth.}
\label{fig:inference_google}
\end{figure}

Table~\ref{tab:google_main} reports results on this dataset. Consistent with its behaviour on WildfireSpreadTS, ShearFuse-UNet outperforms the corresponding TD-FusionUNet baseline at both capacity points, confirming that the Shearlet residual branch provides a reliable and dataset-agnostic improvement. Fig.~\ref{fig:inference_google} shows representative qualitative results
on the Google dataset. Note that parameter counts in this table reflect \emph{total} parameters (including non-trainable Shearlet filter buffers), whereas \cite{luo2025tdfusion} reports trainable parameters only; the two figures therefore differ slightly from those in the reference.

\begin{table}[t]
\centering
\caption{Computational efficiency comparison. Inference time measured
         on a single GPU sample (128$\times$128 input).}
\label{tab:efficiency}
\begin{tabular}{|l|c|c|c|}
\hline
\textbf{Model} & \textbf{GFLOPs} & \textbf{ms/sample} \\
\hline
Base Swin-UNet   & 6.63 & 8.31 \\
\textbf{ShearFuse-UNet} & \textbf{1.35} & \textbf{3.03} \\
\hline
\end{tabular}
\end{table}

\begin{table}[t]
\centering
\caption{Results using ShearFuse-UNet with varying numbers of input days.}
\label{tab:wfts_days}
\renewcommand{\arraystretch}{1.15}
\setlength{\tabcolsep}{6pt}
\begin{tabular}{|c|c|c|c|c|}
\hline
\textbf{Input Days} & \textbf{Parameters} & \textbf{Precision} & \textbf{Recall} & \textbf{F1} \\
\hline
1 & 267k & 0.564 & 0.632 & 0.596 \\
\hline
\textbf{2} & \textbf{269k} & \textbf{0.570} & \textbf{0.634} & \textbf{0.600} \\
\hline
\end{tabular}
\end{table}

\begin{table}[t]
\centering
\caption{Results on the Google Next-Day Wildfire Spread dataset
\cite{huot2022wildfire} using customized preprocessing \cite{luo2025tdfusion}.
Parameter counts are \emph{total} parameters, unlike \cite{luo2025tdfusion} which reports trainable parameters only.}
\label{tab:google_main}
\renewcommand{\arraystretch}{1.15}
\setlength{\tabcolsep}{6pt}
\begin{tabular}{|l|c|c|c|c|}
\hline
\textbf{Model} & \textbf{Params} & \textbf{Precision} & \textbf{Recall} & \textbf{F1} \\
\hline
TD-FusionUNet$^{*}$~\cite{luo2025tdfusion} & 103k & 0.876 & 0.586 & 0.702 \\
\textbf{ShearFuse-UNet$^{*}$}              &  \textbf{67k}  & \textbf{0.876} & \textbf{0.593} & \textbf{0.707} \\
\hline
TD-FusionUNet$^{**}$~\cite{luo2025tdfusion} & 313k & 0.862 & 0.595 & 0.704 \\
\textbf{ShearFuse-UNet$^{**}$}              & \textbf{185k} & \textbf{0.867} & \textbf{0.600} & \textbf{0.709} \\
\hline
\multicolumn{5}{l}{$^{*}$\,convolution base\,=\,4;\quad $^{**}$\,convolution base\,=\,8.}
\end{tabular}
\end{table}

\section{Conclusions}
\label{sec:conclusions}

We proposed ShearFuse-UNet, a lightweight U-Net that implements a structured
spectral attention mechanism inside each encoder block. The three branches: WHT
(Query), DCT (Key), and cone-adapted Shearlet (Value), are functionally
inspired by the division of roles in transformer self-attention, while replacing
learned projections with analytically fixed transforms. The SpectralFusion gate
plays the role of the attention map, producing a data-dependent channel
weighting from the interaction of the two isotropic spectral representations;
the Shearlet residual then injects directionally selective content that neither
the WHT nor the DCT can provide.

This design is particularly well-suited to wildfire spread prediction: the WHT
captures global spatial mixing patterns correlated with broad environmental
drivers, the DCT encodes smoothly varying meteorological and topographic
features, and the Shearlet captures the anisotropic edge geometry of fire
fronts shaped by wind direction and terrain. On WildfireSpreadTS,
ShearFuse-UNet achieves F1\,=\,0.596 with only 267k
parameters, surpassing a 14M-parameter ResNet18-based U-Net and confirming
that structured spectral attention is a principled and highly efficient
inductive bias for geospatial fire spread segmentation.

Results on the Google Next-Day Wildfire Spread dataset \cite{huot2022wildfire} further confirm the dataset-agnostic effectiveness of ShearFuse-UNet. 

Future work will explore end-to-end trainable Shearlet filter banks, explicit
multi-head spectral attention with more than three transform branches, extension
to multi-day time-series prediction, and deployment on edge hardware platforms.
\section*{Acknowledgment}
This work was supported by the National Science Foundation under Award No. 2531376.

\balance
\bibliographystyle{IEEEtran}
\bibliography{main}

\newpage

\begin{IEEEbiographynophoto}{Ene Meco} received the B.S. degree in electrical engineering from Sorbonne University, Paris, France, in 2025. He is currently pursuing the Ph.D.
degree in electrical and computer engineering at University of Illinois at Chicago, IL, USA.
\end{IEEEbiographynophoto}

\begin{IEEEbiographynophoto}{Yingyi Luo} received the B.S. degree in electrical engineering from Northeastern University, Shenyang, liaoning, in 2021 and the M.S. degree in electrical and computer engineering from University of Illinois at Chicago, IL, USA, in 2022. She is currently pursuing the Ph.D.
degree in electrical and computer engineering at University of Illinois at Chicago, IL, USA.
\end{IEEEbiographynophoto}


\begin{IEEEbiographynophoto}{Emadeldeen Hamdan} received the B.S. degree in electrical and computer
engineering from University of Illinois at Chicago, IL, USA, in 2022. He is currently pursuing the Ph.D.
degree in electrical and computer engineering at University of Illinois at Chicago, IL, USA.
\end{IEEEbiographynophoto}

\begin{IEEEbiographynophoto}{Adam Watts} received the M.S. degree in Interdisciplinary Ecology, with a specialization in restoration ecology, from the University of Florida, Gainesville, FL, USA, in 2002, and the Ph.D. degree in Interdisciplinary Ecology, with a specialization in fire ecology, from the same institution in 2012. He is currently a Supervisory Research Biologist and the Fire Environment Research Applications (FERA) Lead with the U.S. Forest Service. Previously, he served as an Associate Research Professor and Deputy Director of the Division of Atmospheric Sciences at the Desert Research Institute, where he also directed the Airborne Systems Testing for Environmental Research (ASTER) Center and led the Division's Unmanned Aircraft Systems (UAS) program. His research interests include wildland fire ecology, ecosystem function and feedbacks, ecological restoration, unmanned aircraft systems for environmental monitoring, and fire management applications.

\end{IEEEbiographynophoto}

\begin{IEEEbiographynophoto}{Ahmet Enis
Cetin} received his Ph.D. in 1987 from the University of Pennsylvania. He was an Assistant Prof. at the University of Toronto between 1987-1989. He was a faculty member at Bilkent University, Ankara, Turkey from 1989 to 2017. Dr Cetin has been a faculty member in the Department of Electrical and Computer Engineering at the University of Illinois at Chicago (UIC) since 2017. He also has held visiting professor positions at Bellcore (1988), University of Minnesota (1996-1997), and UC San Diego (2016-2017). Dr. Cetin has been carrying out research in the areas of theoretical and applied machine learning, signal, image, and video processing, biomedical signal processing, infrared and chemical sensor signal processing in Cyber-Physical Systems (CPS). Erzin and Cetin introduced the concept of adaptive prediction and split vector quantization for Line Spectral Frequency representation. This concept was used in ITU speech coding standards including G.729, G.723.1, and GSM EFR. 

He became a Fellow of IEEE for his contributions to signal and image recovery. In addition, Dr. Cetin is the Editor-in-Chief of Signal, Image and Video Processing at Springer-Nature. He received a best paper award for his camera-based wildfire detection work at a conference organized by UNESCO and the Cyprus Presidency of the European Union.  He is also one of the co-founders of the multinational smart wide-angle OEM camera company Oncam-Grandeye (https://www.oncam.com), UK. Dr. Cetin served as the CEO/CTO of Grandeye, Turkey between 2003-2013. Oncam-Grandeye cameras won design and innovation awards in IFSEC, UK, and ISC WEST, Las Vegas, trade fairs.

\end{IEEEbiographynophoto}

\EOD
\end{document}